\documentclass[runningheads]{llncs}

\usepackage[T1]{fontenc}
\usepackage{amsmath,amssymb}
\usepackage{graphicx}
\usepackage{booktabs}
\usepackage{longtable}
\usepackage[square,numbers,sort&compress]{natbib}
\usepackage{url}
\usepackage[hidelinks]{hyperref}
\usepackage{orcidlink}

\newcommand{\resourcename}{DesignABO}
\newcommand{\expandedsetting}{Expanded Setting}
\newcommand{\structuresetting}{Structure-Sensitive Setting}
\newcommand{\humanprotocol}{Human-Graded Reference Protocol}

\title{Sketch2Inspire: Structure-Sensitive Evaluation for Product Retrieval}
\titlerunning{Sketch2Inspire}

\author{Ge Kong\,\orcidlink{0009-0007-5486-8605}}
\authorrunning{Ge Kong}
\institute{Beihang University\\\email{gekong@buaa.edu.cn}}

\begin{document}

\maketitle

\begin{abstract}
Early-stage product design retrieval often requires more than category recognition: designers may need reference examples that match both a short semantic intent and a rough structural cue. Existing product-image resources and generic image--text retrieval benchmarks rarely separate category retrieval from within-category structural fit. We present Sketch2Inspire, built from a curated subset of Amazon Berkeley Objects with aligned text queries, edge-based sketch-proxy queries, and fused text--sketch queries. The resource separates broad category-level retrieval from structure-sensitive within-category retrieval and includes a human-graded reference protocol for calibration. We evaluate a lightweight reference system based on pretrained CLIP-family encoders, comparing text-only retrieval, sketch-only retrieval, weighted late fusion, and text-first reranking without updating model weights. Under broad relevance, late fusion obtains the highest score (nDCG = 0.9962). Under automatic structure-sensitive relevance, late fusion again obtains the highest score (nDCG = 0.7015), exceeding text-only retrieval (nDCG = 0.5912). In the human-graded results, late fusion obtains the highest nDCG@10 (0.9133), while text-only retrieval ranks second (0.9030). These results show that the retrieval gain from multimodal input depends on how relevance is defined. Sketch2Inspire therefore provides a diagnostic resource for evaluating modality contribution and supports the development of structure-aware product-retrieval protocols with independent human annotation.

\end{abstract}

\section{Introduction}

Designers often search for visual references before a concept is fully specified. In that stage, useful retrieval should not only recover the correct product category but also surface shapes, silhouettes, proportions, and visual analogies that can guide ideation~\citep{westerman2007creativeproduct,kwon2021multimodaldesign,kwon2022enabling,kwon2023understanding}. This makes product-design inspiration search different from generic image retrieval, where category recognition is often a sufficient proxy for success.

Current resources only partly support this setting. Product datasets such as Amazon Berkeley Objects (ABO) and fashion-oriented retrieval corpora provide rich images and metadata, but they are not organized as aligned sketch--text--product evaluation resources for design search~\citep{collins2022abo,rostamzadeh2018fashiongen,wu2021fashioniq,gao2020fashionbert,goenka2022fashionvlp,wang2023fashionklip,islam2024fashionsurvey}. Recent sketch+text retrieval datasets in other domains point in a related direction~\citep{xu2026thangka}, but they do not target product-design inspiration. More importantly, many retrieval protocols still score all same-category matches similarly, making it hard to determine whether a sketch cue improves the ranking or whether text alone already captures most of what is measured.

Sketch2Inspire addresses this evaluation gap with a traceable resource. We construct \resourcename{} from a curated ABO subset and define two complementary views of the task: \expandedsetting{} for broad category-level retrieval and \structuresetting{} for within-category structural disambiguation. We additionally report \humanprotocol{}, which provides a manual check separate from the automatic relevance rules. Our study asks when sketch cues change product-retrieval rankings in ways that matter for design support, and explores a retrieval scheme that combines textual and visual evidence.

The reference retrieval system uses pretrained CLIP-family encoders to compute text--image and sketch--image similarity separately, and then compares text-only retrieval, sketch-only retrieval, weighted late fusion, and text-first reranking. This design supports attribution: changes in ranking can be interpreted through the query modality and the relevance definition, rather than through changes in model training.

The study is scoped to controlled ranking evaluation, attribution of modality contribution, and evaluation-protocol design. Claims about real design-support systems, designer preference, and learned retrieval architectures require separate user-centred and methodological studies. The role of the resource is to reveal where multimodal scores change a ranking and where category-level text matching already explains the measured performance.

\begin{figure}[t]
\centering
\includegraphics[width=\textwidth]{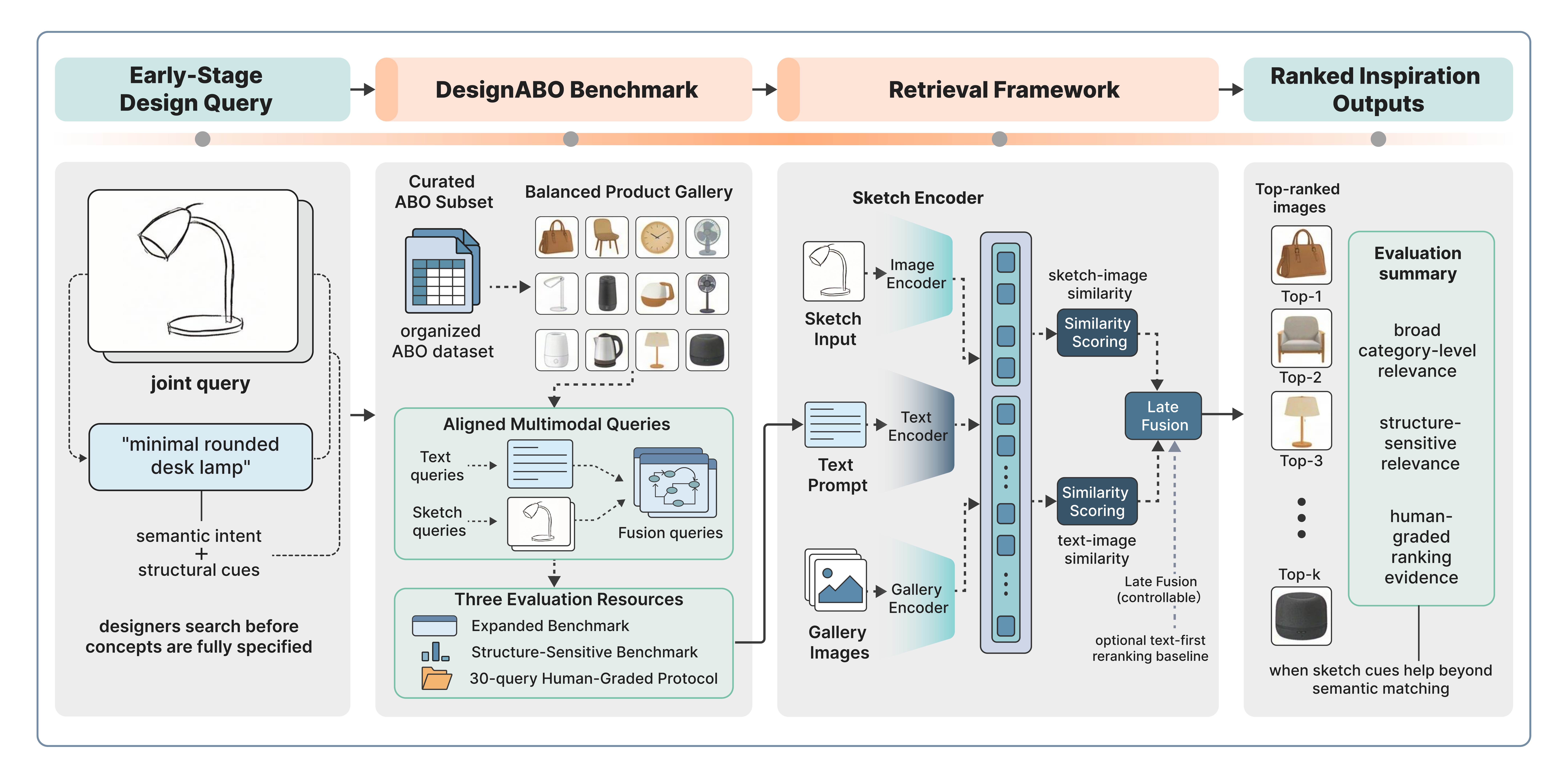}
\caption{Sketch2Inspire overview. The evaluation resource pairs controlled text and sketch-proxy queries with a curated product gallery, then evaluates text-only, sketch-only, late-fusion, and reranking retrieval under broad category-level, structure-sensitive, and human-graded protocols.}
\label{fig:overview}
\end{figure}

This framing leads to four specific contributions.

\textbf{Design-oriented evaluation resource.} We curate \resourcename{}, a traceable product-image retrieval resource with aligned multimodal queries and a structure-sensitive subset.

\textbf{Evaluation protocol.} We explicitly separate category-level relevance from structure-sensitive relevance and report a human-graded reference protocol, making the limitations of broad same-category scoring visible.

\textbf{Controlled reference system.} We use a transparent zero-shot retrieval framework based on pretrained CLIP-family embeddings, late fusion, and text-first reranking. The system serves as an interpretable probe for isolating modality contribution.

\textbf{Empirical finding.} Late fusion obtains the highest nDCG under broad relevance, automatic structure-sensitive relevance, and the human-graded protocol, while the size and interpretation of the gain depend on the relevance definition.

\section{Related Work}

\paragraph{Product and design retrieval.}
ABO provides a large collection of product catalog images, metadata, and 3D assets for real-world object understanding~\citep{collins2022abo}. Fashion-oriented datasets and benchmarks have further advanced image--text retrieval, natural-language feedback, and e-commerce search~\citep{rostamzadeh2018fashiongen,wu2021fashioniq,gao2020fashionbert,goenka2022fashionvlp,wang2023fashionklip,islam2024fashionsurvey}. These resources are valuable, but they are not built around early-stage product-design search, where rough visual intent and inspiration-oriented matching are central. Prior design-search studies show that examples, visual analogies, and multimodal search interfaces can affect how designers discover inspiration~\citep{westerman2007creativeproduct,kwon2021multimodaldesign,kwon2022enabling,kwon2023understanding,bako2022datavizexamples}.

\paragraph{Vision--language and composed retrieval.}
Cross-modal retrieval has benefited from shared representation learning in CLIP, ALBEF, BLIP, BLIP-2, ViLT, and related retrieval architectures~\citep{radford2021clip,li2021albef,li2022blip,li2023blip2,kim2021vilt,zhang2020caan,rao2023retrievalaugvlp}. Composed and feedback-based retrieval methods combine an image-like signal with natural language supervision~\citep{vo2019tirg,anwaar2021compositional,chen2020visiolinguistic,dodds2020maaf}. Recent sketch-driven work studies sketch--text duets, sketch abstraction, diffusion-based sketch--photo matching, and training-free reasoning for composed retrieval~\citep{koley2024duet,koley2024abstraction,koley2024matchmakers,sun2025cotmr}. Sketch2Inspire is complementary to these methods because it provides a controlled evaluation setting that makes category-level and structure-sensitive retrieval behavior separable.

The present experiments use CLIP-family encoders and score-level fusion as a transparent reference system. Each retrieval mode uses the same gallery embeddings, query set, and relevance protocol, so changes in ranking can be attributed to the availability of text, sketch, or fused evidence. Stronger composed-retrieval architectures, including BLIP-style retrieval, diffusion-assisted sketch--photo matching, and larger multimodal backbones, remain important comparisons for absolute performance. The present study isolates a different issue: whether an evaluation protocol can show that a sketch cue changes ranking behavior beyond category-level text matching.

\section{Evaluation Resource and Reference Retrieval}

\subsection{ABO Subset and Query Construction}

\resourcename{} is built from a curated subset of ABO product images. \expandedsetting{} contains 9000 product images from 300 categories, with 30 images per category; examples include bag, chair, clock, fan, humidifier, kettle, lamp, and speaker. The categories were chosen because they are common product-design targets and because within-category silhouette differences can plausibly matter for design. The gallery is tied to fixed file lists rather than random online retrieval, which makes the reported rankings traceable and repeatable.

\structuresetting{} retains 3900 product images from 260 categories, with 15 images per category; examples include bag, chair, clock, kettle, lamp, and speaker. Fan and humidifier were removed because their available examples in the curated subset were less informative for silhouette-driven structural disambiguation. Structural differences are defined operationally as within-category variation in silhouette, aspect ratio, dominant part layout, foreground occupancy, and contour complexity. Within each retained category, near duplicates, category outliers, background-dominant product shots, and examples with low structural contrast were filtered using within-category embedding density, silhouette similarity, and simple visual statistics. This produces a focused evaluation gallery in which shape, proportion, and contour differences are more consequential than in the broad setting. The curation is deterministic and reproducible from the fixed file lists, but it is not presented as evidence that the task difficulty has been independently validated by designers. \humanprotocol{} complements these automatic settings with 300 manually graded query--image pairs.

To make the resource auditable, each retained item is represented by a fixed record containing its image identifier, category, source identifier, generated sketch path, selection score, and low-level visual statistics such as edge density, foreground area ratio, aspect ratio, and bounding-box fill. Each query record stores the text prompt, category, anchor image, excluded anchor identifier, and the relevance lists used by the automatic protocols. These records define the gallery, query set, and evaluation labels used in the reported experiments, rather than relying on ad hoc sampling at evaluation time.

For controlled evaluation, each sketch query is an edge-based sketch proxy generated from an anchor product image. Product images are resized with white padding to $256 \times 256$, converted to grayscale, bilaterally filtered, processed with Canny thresholds of 40 and 120, dilated with a $2 \times 2$ kernel, and inverted to produce black contours on a white background. This design makes the structural cue reproducible and keeps text, sketch, and fusion queries aligned. It also means that the present study evaluates controlled structural proxies rather than the full variability of real hand-drawn ideation sketches. To reduce direct self-match leakage, the anchor identifier is stored in the query record and the anchor product is removed from all candidate pools, relevance lists, and rankings before evaluation.

Each gallery item $i \in \mathcal{G}$ has a product image $x_i$ and category label $c_i$. A sketch-proxy query is generated by an edge operator $g(\cdot)$:
\begin{equation}
x_q^{(s)} = g(x_{a(q)}),
\end{equation}
where $a(q)$ is the anchor image for query $q$. A paired text query $t_q$ describes the intended product type and style at a short-prompt level. The paired design intentionally leaves some structural information to the sketch branch: the text prompt anchors semantic intent, while the sketch proxy supplies a controlled silhouette cue.

\subsection{Reference Retrieval Modes}

Let $f_{\mathrm{img}}(\cdot)$ and $f_{\mathrm{txt}}(\cdot)$ denote the image and text encoders of a pretrained CLIP-family model~\citep{radford2021clip}. Gallery images, sketch proxies, and text queries are normalized as
\begin{equation}
z_i = \frac{f_{\mathrm{img}}(x_i)}{\|f_{\mathrm{img}}(x_i)\|_2}, \quad
z_q^{(s)} = \frac{f_{\mathrm{img}}(x_q^{(s)})}{\|f_{\mathrm{img}}(x_q^{(s)})\|_2}, \quad
z_q^{(t)} = \frac{f_{\mathrm{txt}}(t_q)}{\|f_{\mathrm{txt}}(t_q)\|_2}.
\end{equation}
The modality-specific scores are
\begin{equation}
s_{\mathrm{text}}(q,i) = \langle z_q^{(t)}, z_i \rangle, \qquad
s_{\mathrm{sketch}}(q,i) = \langle z_q^{(s)}, z_i \rangle.
\end{equation}

We compare four retrieval modes. Text-only ranks by $s_{\mathrm{text}}$, sketch-only ranks by $s_{\mathrm{sketch}}$, and late fusion ranks by
\begin{equation}
s_{\mathrm{fusion}}(q,i) = \alpha s_{\mathrm{sketch}}(q,i) + (1-\alpha) s_{\mathrm{text}}(q,i).
\end{equation}
The text-first reranking baseline first keeps the top-$M$ text candidates and then reranks them with sketch similarity:
\begin{equation}
\mathcal{C}_q^{(M)} = \operatorname{TopM}_{i \in \mathcal{G}} s_{\mathrm{text}}(q,i),
\qquad
\pi_{\mathrm{rerank}}(q) = \operatorname{argsort}_{i \in \mathcal{C}_q^{(M)}} s_{\mathrm{sketch}}(q,i),
\end{equation}
with $M=20$ in the reported experiments.

All reported embeddings use the MobileCLIP-S1 checkpoint pretrained with \texttt{datacompdr}~\citep{vasu2024mobileclip,gadre2023datacomp}. We use batch size 64 for the automatic experiments. No model weights are updated. The automatic retrieval runs use cosine similarity over $\ell_2$-normalized embeddings. Late-fusion weights are evaluated on the grid reported in Table~\ref{tab:alpha}; the automatic structure-sensitive table reports the nDCG-maximizing setting from that grid, whereas the human-graded table reports the fixed $\alpha=0.60$ fusion setting used to construct the fixed manual candidate pool.

The retrieval framework is deliberately model-light, so that modality effects can be inspected over pretrained representations. The experiments include multiple retrieval modes over the same embeddings: the comparison isolates whether sketch information changes rankings under different relevance definitions. These modes function as interpretable probes of modality contribution, which also makes negative results informative: when text-only wins, the resource indicates that the relevance protocol may be dominated by semantics rather than structure.

\subsection{Relevance Protocols}

The broad protocol treats non-anchor items from the same category as relevant:
\begin{equation}
\mathcal{R}_q^{\mathrm{broad}} = \{ i \in \mathcal{G} \mid c_i = c_{a(q)}, i \neq a(q) \}.
\end{equation}
The automatic structure-sensitive protocol keeps a focused within-category set closest to the anchor under a shape-weighted relevance score. Let $u_i$ denote a normalized silhouette descriptor and $v_i$ denote a normalized contour descriptor for gallery item $i$. The silhouette descriptor is a downsampled $64 \times 64$ foreground mask. The contour descriptor contains aspect ratio, foreground area, bounding-box fill, circularity, edge density, and contour-area ratio. The structure-sensitive score is
\begin{equation}
r_{\mathrm{struct}}(q,i) =
0.30\langle z_{a(q)}, z_i\rangle +
0.55\langle u_{a(q)}, u_i\rangle +
0.15\langle v_{a(q)}, v_i\rangle .
\end{equation}
The high-relevance set is then defined as
\begin{equation}
\mathcal{R}_q^{\mathrm{strict}} =
\operatorname{TopK}_{i \in \mathcal{R}_q^{\mathrm{broad}}}
r_{\mathrm{struct}}(q,i), \qquad K=5.
\end{equation}
This rule is useful as a controlled diagnostic setting, but it is not an independent ground-truth annotation. One component uses the same CLIP-family image space as the reference retriever, and the silhouette and contour terms remain heuristic. For that reason, we interpret automatic structure-sensitive gains as evidence of a controlled modality effect and use the human-graded protocol to calibrate the main claim.

\section{Experiments}

\subsection{Settings and Metrics}

Both automatic settings use aligned text, sketch-proxy, and fusion query sets so that retrieval modes are compared within their respective fixed galleries. Table~\ref{tab:resources} summarizes the automatic evaluation resources. We report Recall@5, Recall@10, mAP, and nDCG for the automatic settings, and nDCG@10, mAP@10, and Recall@5 for the human-graded protocol.

The human-graded pool is treated as a reference check rather than a full-gallery annotation. For each fusion query, candidate images were pooled from the top-ranked outputs of text-only, sketch-only, and late-fusion retrieval. Duplicate images were merged, and the final candidates were selected by reciprocal-rank-fusion score, number of contributing methods, best observed rank, fusion rank, and image identifier. Each query--image pair was then assigned a final relevance label in $\{2,1,0\}$. A label of 2 indicates that the result matches the intended product category and preserves the main structural cue in the sketch proxy; a label of 1 indicates either a plausible category match with weaker structural fit or a structurally plausible reference with weaker semantic fit; a label of 0 indicates that the result is not useful for the query. The grading task therefore concerns query--image relevance, not designer preference, creative usefulness, or interaction quality. This design keeps methods comparable because all methods are evaluated over the same graded candidate set, but it narrows the conclusion to ranking quality inside a curated candidate pool. The archived annotation file records only the query identifier, image identifier, and final grade; it does not record annotator identity, design-expertise metadata, duplicate ratings, or disagreement resolution. We therefore report the protocol as a preliminary human-graded reference and do not estimate inter-rater agreement.

\begin{table}[t]
\centering
\small
\caption{Evaluation resources used in Sketch2Inspire. The human-graded protocol is reported separately from the automatic structure-sensitive setting.}
\label{tab:resources}
\begin{tabular}{lcccc}
\toprule
Setting & Cats. & Img./cat. & Total & Query coverage \\
\midrule
\expandedsetting{} & 300 & 30 & 9000 & text / sketch / fusion \\
\structuresetting{} & 260 & 15 & 3900 & text / sketch / fusion \\
\bottomrule
\end{tabular}
\end{table}

\subsection{Broad Category-Level Retrieval}

Table~\ref{tab:expanded} shows that late fusion achieves the highest nDCG on \expandedsetting{}. Under broad category-level relevance, late fusion reaches nDCG = 0.9962, while text-only retrieval remains a strong baseline with Recall@10 = 0.4983 and mAP = 0.9839. This result indicates that when all same-category images are treated as relevant, semantic text cues explain much of the task, but the fused query can still improve top-rank ordering.

This result provides a coarse-relevance baseline. It shows that a product-design retrieval task can appear nearly solved under broad same-category scoring even when it has not tested the structural aspect of design search. In other words, the high mAP and nDCG values should not be read as evidence that the system has solved inspiration retrieval. They mostly show that pretrained vision--language representations cluster product categories well enough for coarse relevance.

\begin{table}[t]
\centering
\small
\caption{Automatic results on \expandedsetting{} under broad category-level relevance.}
\label{tab:expanded}
\begin{tabular}{lcccc}
\toprule
Method & Recall@5 & Recall@10 & mAP & nDCG \\
\midrule
text-only & 0.2500 & 0.4983 & 0.9839 & 0.9698 \\
sketch-only & 0.2117 & 0.3983 & 0.7686 & 0.8984 \\
late fusion ($\alpha = 0.6$) & 0.2417 & 0.4683 & 0.9047 & 0.9962 \\
\bottomrule
\end{tabular}
\end{table}

\subsection{Structure-Sensitive Retrieval}

Table~\ref{tab:structure} shows a different pattern under \structuresetting{}. For category-level relevance, text-only remains highly competitive. For fine-grained structural relevance, late fusion reaches the highest nDCG and mAP, with Recall@10 = 0.9000, mAP = 0.5290, and nDCG = 0.7015. Text-first reranking is also competitive, which supports the interpretation that sketch cues are most useful as structural refinement after semantic anchoring. The reranking baseline is included to test whether sketch information is more effective after text has first constrained the semantic candidate pool.

The contrast between the two rows of Table~\ref{tab:structure} is the central diagnostic result. Under category-level relevance, adding sketch information does not materially improve over text-only retrieval. Under fine-grained structural relevance, sketch-only retrieval improves over text-only on Recall@5 and nDCG, and late fusion gives the highest overall ranking metrics. This pattern suggests that the value of sketch input is evaluation-dependent: it becomes visible only when the relevance protocol rewards within-category structure rather than category membership alone.

\begin{table}[t]
\centering
\small
\caption{Automatic results on \structuresetting{}. The late-fusion row uses the highest-scoring fine-grained structural setting ($\alpha = 0.40$), and text-first reranking uses a text candidate pool of 20 images.}
\label{tab:structure}
\begin{tabular}{llcccc}
\toprule
Setting & Method & Recall@5 & Recall@10 & mAP & nDCG \\
\midrule
category-level & text-only & 0.3500 & 0.6881 & 0.9606 & 0.9866 \\
category-level & sketch-only & 0.3048 & 0.5810 & 0.8132 & 0.9117 \\
category-level & late fusion & 0.3548 & 0.6810 & 0.9548 & 0.9858 \\
category-level & text-first reranking & 0.3452 & 0.6571 & 0.9160 & 0.9581 \\
\midrule
fine-grained structural & text-only & 0.3778 & 0.7667 & 0.3902 & 0.5912 \\
fine-grained structural & sketch-only & 0.5889 & 0.7333 & 0.5070 & 0.6785 \\
fine-grained structural & late fusion & 0.6111 & 0.9000 & 0.5290 & 0.7015 \\
fine-grained structural & text-first reranking & 0.6000 & 0.8333 & 0.5177 & 0.6860 \\
\bottomrule
\end{tabular}
\end{table}

The alpha sweep in Table~\ref{tab:alpha} further supports this reading. The highest-scoring late-fusion point lies in the middle range rather than at a single-modality extreme. Performance peaks around $\alpha=0.40$ under fine-grained structural relevance, indicating that the sketch branch helps most when it complements rather than replaces the text branch. This ablation is therefore not presented as a learned weighting strategy; it is a controlled probe of how much structural evidence is useful under a given relevance definition.

\begin{table}[t]
\centering
\small
\caption{Alpha sweep on \structuresetting{} under fine-grained structural relevance.}
\label{tab:alpha}
\begin{tabular}{lcccc}
\toprule
Variant & Recall@5 & Recall@10 & mAP & nDCG \\
\midrule
$\alpha = 0.05$ & 0.4111 & 0.8000 & 0.4092 & 0.6035 \\
$\alpha = 0.10$ & 0.4556 & 0.8444 & 0.4374 & 0.6276 \\
$\alpha = 0.20$ & 0.5111 & 0.9000 & 0.4519 & 0.6333 \\
$\alpha = 0.30$ & 0.5778 & 0.9222 & 0.4821 & 0.6613 \\
$\alpha = 0.40$ & 0.6111 & 0.9000 & 0.5290 & 0.7015 \\
$\alpha = 0.50$ & 0.6222 & 0.8778 & 0.5252 & 0.6962 \\
$\alpha = 0.60$ & 0.6000 & 0.8333 & 0.5083 & 0.6822 \\
$\alpha = 0.80$ & 0.6111 & 0.7778 & 0.5286 & 0.6995 \\
text-first reranking & 0.6000 & 0.8333 & 0.5177 & 0.6860 \\
\bottomrule
\end{tabular}
\end{table}

\subsection{Human-Graded Evidence}

Table~\ref{tab:human} reports the completed human-graded results. This protocol provides a manual calibration separate from the automatic structure-sensitive setting. Late fusion achieves the highest manual nDCG@10, reaching 0.9133, while text-only retrieval ranks second on nDCG@10 at 0.9030 and remains strong on mAP@10. This result is central to the paper's interpretation: fusion can improve ranking-sensitive relevance under the manual protocol, but the size and meaning of the gain still depend on the relevance definition.

The human-graded result also clarifies the paper's contribution as an evaluation study. The same retrieval system can support different conclusions depending on whether relevance is broad, automatically structure-sensitive, or manually graded. The manual grades support the usefulness of separating structure-aware evaluation from category-level retrieval, while identifying stronger human validation as a requirement for robust fusion claims.

\begin{table}[t]
\centering
\small
\caption{Results on \humanprotocol{}. This human-graded evaluation is reported as a separate manual check and is not part of the automatic structure-sensitive protocol.}
\label{tab:human}
\begin{tabular}{lccc}
\toprule
Method & manual nDCG@10 & manual mAP@10 & manual Recall@5 \\
\midrule
text-only & 0.9030 & 0.9479 & 0.5336 \\
sketch-only & 0.8888 & 0.8945 & 0.4722 \\
late fusion & 0.9133 & 0.9196 & 0.5206 \\
\bottomrule
\end{tabular}
\end{table}

\subsection{Qualitative Cases}

\begin{figure}[t]
\centering
\includegraphics[width=\textwidth]{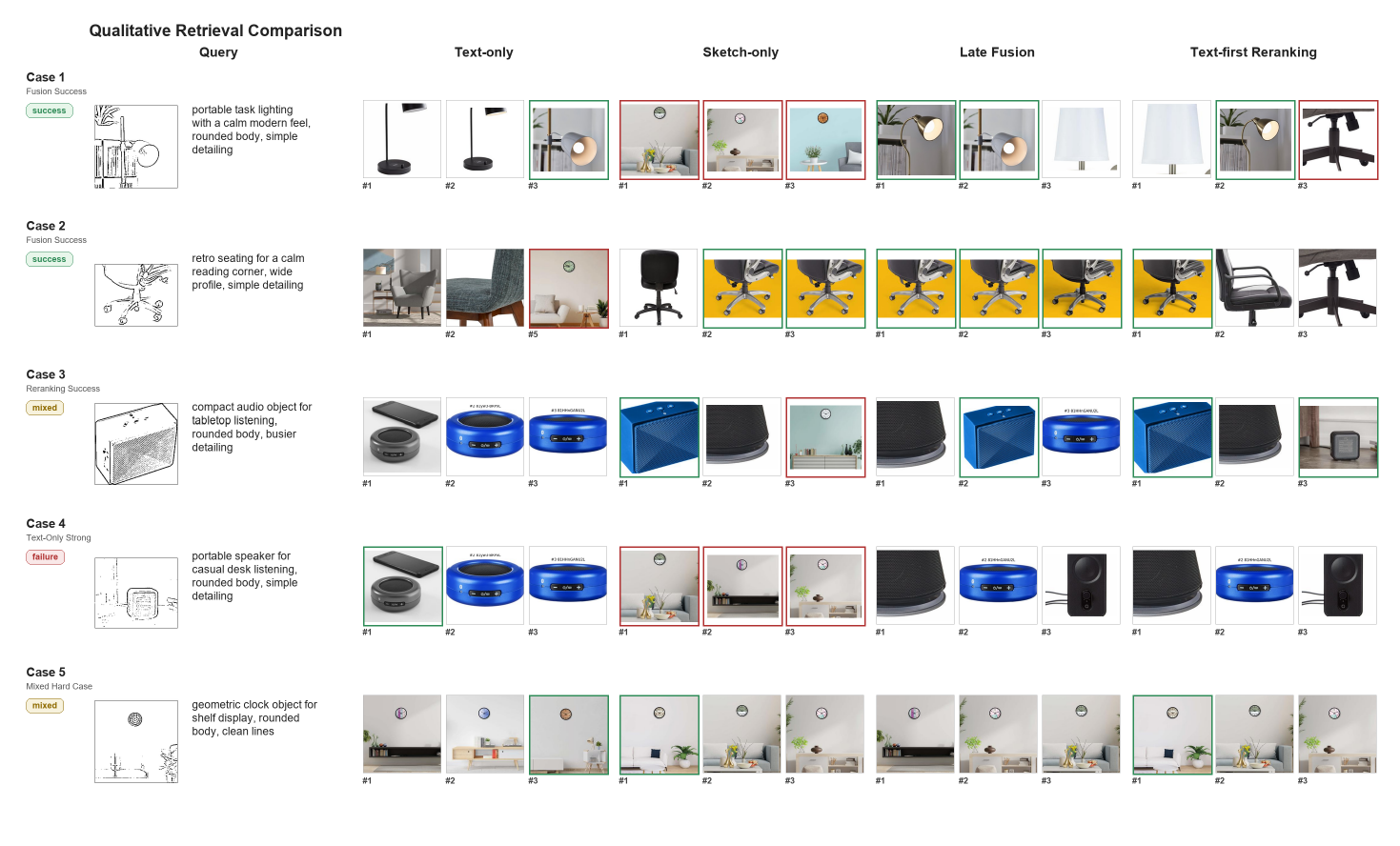}
\caption{Qualitative retrieval comparison on representative cases from \structuresetting{}. Each row shows a sketch-proxy query and text prompt, followed by top retrieved examples from text-only retrieval, sketch-only retrieval, late fusion, and text-first reranking. Borders indicate stored proxy annotations used for visualization. The figure illustrates why text can dominate broad category matching while sketch cues can help with within-category structural drift.}
\label{fig:qualitative}
\end{figure}

The qualitative cases in Figure~\ref{fig:qualitative} make the aggregate pattern more concrete. Text-only retrieval can return semantically plausible products while missing the intended silhouette or part structure. Sketch-only retrieval can recover shape-consistent references, but it may also lose category control when the sketch proxy is ambiguous. Text-first reranking is useful in cases where the text prompt narrows the semantic candidate pool and the sketch cue then reorders structurally relevant alternatives.

The same cases also show why the results should not be read as evidence of a general improvement from fusion. When the text prompt already captures the design intent, adding a sketch signal can introduce noise or over-emphasize low-level contours. A structure-aware evaluation should therefore expose both successful shape correction and text-dominant failures, because both are relevant to design-support retrieval.

\section{Discussion}

The experiments identify a consistent boundary on how sketch cues matter. \expandedsetting{} shows that late fusion can obtain the highest nDCG under broad category-level relevance, while text-only retrieval remains a strong coarse baseline. \structuresetting{} shows a more diagnostic pattern: when relevance depends on within-category structural fit, sketch cues improve the ranking and late fusion achieves the highest automatic nDCG and mAP. The \humanprotocol{} also places late fusion first on nDCG@10, while the small margin over text-only and the accompanying metrics show that the gain should be interpreted with respect to the relevance definition that produced it. Taken together, these results support structure-sensitive evaluation as a necessary complement to category-level product retrieval metrics.

This distinction is important for multimodal retrieval evaluation. If an evaluation protocol treats all same-category results as equivalent, then a system can achieve high scores by recovering the product category, even when the sketch does not influence the ranking in a design-relevant way. Conversely, fine-grained automatic labels can reflect assumptions in the representation used to construct them, especially when the same model family also supplies the retrieval embeddings. Sketch2Inspire addresses this tension by reporting category-level, structure-sensitive, and human-graded evidence side by side. The resulting disagreements help characterize the role of each modality and define where further validation is needed.

The contribution of Sketch2Inspire is an evaluation resource and analysis framework for measuring modality contribution under controlled relevance definitions. The resource is intentionally scoped to make retrieval behavior traceable: it aligns text, sketch-proxy, and fusion queries over fixed galleries and evaluates them under progressively stricter relevance protocols. Its role is therefore diagnostic rather than definitive. The study asks a specific evaluation question: when does multimodal input improve ranking beyond category-level text matching, and under what conditions does the sketch signal provide design-relevant structure?

\section{Limitations}

\textbf{Sketch realism.} The sketch inputs are edge-based proxies generated from product photos rather than real early-stage sketches from designers. Anchor exclusion prevents the original product image from being returned as its own match, but the proxy sketch still remains more visually coupled to product photographs than a freehand ideation sketch would be. This improves control and reproducibility but may overestimate alignment between query sketches and gallery images. It also limits claims about roughness, ambiguity, and abstraction in real ideation sketches. Future versions should collect real hand-drawn sketches and test robustness to abstraction level and ambiguity.

\textbf{Automatic relevance coupling.} The structure-sensitive automatic protocol uses a weighted heuristic that includes CLIP image similarity, silhouette similarity, and contour similarity. The CLIP component is coupled to the model family used by the reference retriever, while the shape descriptors are still hand-designed proxies. This makes the automatic protocol a controlled diagnostic setting rather than independent ground truth. The human-graded protocol partially addresses this issue, but it remains limited to a shared candidate pool.

\textbf{Method scope.} The retrieval method is a transparent zero-shot reference system designed to isolate modality contribution. Stronger composed-retrieval models, diffusion-assisted sketch--photo matching, and larger-scale multimodal backbones remain necessary for evaluating absolute retrieval performance. Consequently, the reported scores support claims about modality contribution within this evaluation setting, not claims about the competitiveness of late fusion against those methods. Those models should be added in future comparisons using the same category-level, structure-sensitive, and human-graded protocols.

\section{Conclusion}

This study set out to examine when sketch input contributes beyond text-only category matching in product-inspiration retrieval. Using fixed CLIP-family embeddings, the experiments showed that late fusion achieved the highest nDCG under broad category-level relevance and the highest automatic scores when relevance rewarded within-category structure. The human-graded protocol also gave the highest nDCG@10 to late fusion and placed text-only retrieval second, indicating that the fusion gain is visible under ranking-sensitive relevance but should still be interpreted as relevance-dependent rather than as a universal retrieval improvement. Thus, Sketch2Inspire contributes a controlled diagnostic setting for measuring modality contribution under explicit relevance definitions. Because the current galleries are controlled and the sketch inputs are edge-based proxies, the results should be treated as a foundation for larger resources with real designer sketches, independent relevance annotation, broader baselines, and user-centred validation of design-inspiration usefulness.

\bibliographystyle{splncs04}
\bibliography{acml26}

\end{document}